# DESIGN OF A BED ROTATION MECHANISM TO FACILITATE IN-SITU PHOTOGRAMMETRIC RECONSTRUCTION OF PRINTED PARTS

**Travis A. Roberts**
PhD Student
Clemson University
Mechanical Engineering
Fluor Daniel EIB
Clemson, SC
robert2@g.clemson.edu

**Sourabh Karmakar**
PhD Student
Clemson University
Mechanical Engineering
Fluor Daniel EIB
Clemson, SC
skarmak@clemson.edu

**Cameron J. Turner**
Associate Professor
Clemson University
Mechanical Engineering
206 Fluor Daniel EIB
Clemson, SC
cturne9@clemson.edu

**Abstract**

*Additive manufacturing, or 3D printing, is a complex process that creates free-form geometric objects by sequentially placing material to construct an object, usually in a layer-by-layer process. One of the most widely used methods is Fused Deposition Modeling (FDM). FDM is used in many of the consumer-grade polymer 3D printers available today. While consumer grade machines are cheap and plentiful, they lack many of the features desired in a machine used for research purposes and are often closed-source platforms. Commercial-grade models are more expensive and are also usually closed-source platforms that do not offer flexibility for modifications often needed for research. The authors designed and fabricated a machine to be used as a test bed for research in the field of polymer FDM processes. The goal was to create a platform that tightly controls and/or monitors the FDM build parameters so that experiments can be repeated with a known accuracy. The platform offers closed loop position feedback, control of the hot end and bed temperature, and monitoring of environment temperature and humidity. Additionally, the platform is equipped with cameras and a mechanism for in-situ photogrammetry, creating a geometric record of the printing throughout the printing process. Through photogrammetry, backtracking and linking process parameters to observable geometric defects can be achieved. This paper focuses on the design of a novel mechanism for spinning the heated bed to allow for photogrammetric reconstruction of the printed part using a minimal number of cameras, as implemented on this platform.*

## 1. INTRODUCTION

Traditional manufacturing methods, such as machining, molding, and forging, use process qualification techniques to certify part quality because they are significantly faster and cheaper than certifying individual parts. This is only possible because the physics and mechanics of these processes are understood well enough to believe that control of process parameters will produce the expected results. This is currently not the case for additive manufacturing and no additive manufacturing processes are currently qualified to produce parts for aerospace or defense applications [1]. This demonstrates a clear knowledge gap that needs to be explored and overcome to progress this technology.

Process feedback for machine tools gives the ability to monitor and adjust machine parameters during the fabrication process and decreases the number of defective parts. This principle has been applied prolifically to subtractive machining, but limited research has been done on its application to additive manufacturing systems. Additionally, process feedback provides the ability to certify a part without slow and expensive after-process inspections. However, each part currently needs to be qualified using evaluation methods that require additional skilled personnel before it can be put into service in a critical application. Process feedback can change this by certifying a part as it is being made, reducing wasted time and material by mitigating defects. The NIST (National Institute of Standards and Technology) Roadmap for Additive Manufacturing lists development of real-time process monitoring techniques and feedback systems as key goals for the advancement of additive manufacturing techniques [2].

While many of the faults, or defects, in subtractive machined parts are results of problems with the machine, such as failing bearings, tool wear, chatter and breakage, many of the faults in additive machining are based around material errors that can be traced back to printing parameters [3]–[7], such as build-plate separation, delamination, and dimensional changes due to thermal distortion. These faults can be placed into two categories: tolerable faults and terminal faults. Tolerable faults are deviations in the printed part from the ideal model that do not render the artifact unusable; dimensional tolerances, surface profile, and material density fall into this category. Tolerable faults may need rework to be put into service. Terminal faults are deviations that scrap the printed part such as delamination, positioning problems, and build-plate separation. Terminal faults are a source of inefficiency in the additive manufacturing process that result in loss of machine time and wasted materials. In-situ monitoring of artifacts during the manufacturing process is required to detect these faults in time for corrective action to be taken and turn terminal faults into tolerable ones.

Several of the faults present in FDM printing process, such as delamination, build-plate separation, and some dimensional changes, stem from the material changes that occur during the printing process. FDM printing works by heating a filament of material until it can be extruded into a specific position as the print head moves in space. The print head extrudes layer upon layer of material to build the artifact in the desired shape. During this process, the material goes through a heating and cooling cycle related to the print head depositing new layers of heated filament, as shown by [1]. [2] showed that the cooling profile during this process results in residual thermal stresses and strains, with tensile stresses on the top of the part and compressive stresses on the bottom. [3]–[5] showed that the resulting distortions can be reduced by decreasing layer height, object. This paper focuses on the design of a novel mechanism that allows the machine to take pictures of printed objects in-decreasing heat input, preheating the build-plate, and insulating the part to control the cooling profile.

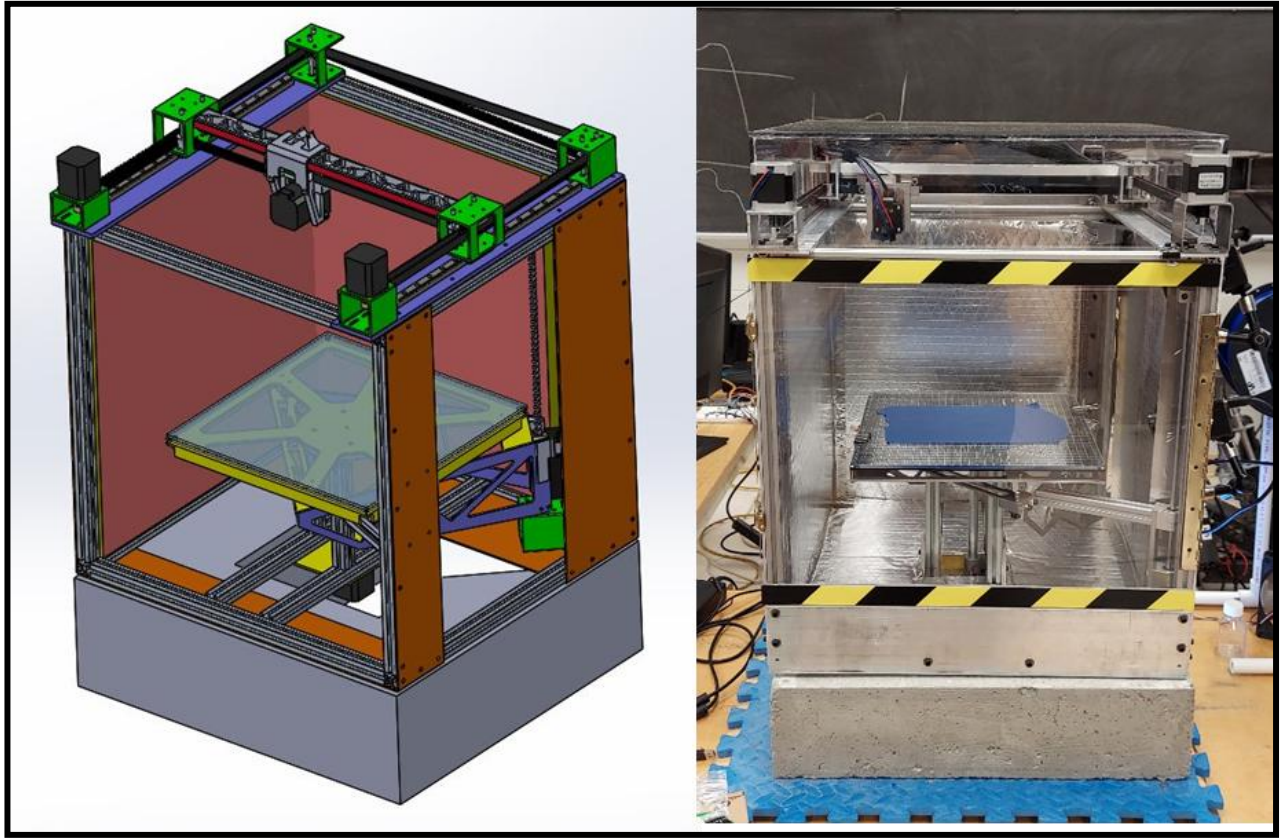

**Figure 1**. CAD model and as-built FDM additive manufacturing platform.

The authors of this paper are interested in linking these defects to the building parameters that birthed them. To do this, direct knowledge of the printing parameters is desired. However, in addition to the parameter abstraction introduced by slicing software, there can also be significant differences between what the printer is commanded to do and what it actually achieves. To further this line of investigation, the authors designed and constructed an FDM additive manufacturing platform that can record machine parameters during the build process. A picture of the system can be viewed in Figure 1. The printer uses a CoreXY motion architecture and has a build volume of 300 mm x 300 mm x 265 mm. All axes use linear rails for motion. The direct extruder is water cooled (Titan Aqua by E3D) for improved performance in the enclosed environment. The heated bed is a 1200W 120VAC model from E3D. The platform is equipped with various sensors to measure different build parameters during the printing process including print head and bed temperature, axis position and speed, and environmental temperature and humidity.

One of the desires for this platform was to have a system for in-situ capture and detection of defects. Photogrammetry is the process of using pictures of an object taken from various viewpoints to reconstruct 3D geometric information about the situ during the printing process and the supporting system to use those pictures for photogrammetric reconstruction.

The next section will present the state of the art in-situ part monitoring in additive manufacturing and kinematic couplings. Section 3 will detail the design of the bed rotation mechanism and section 4 will give details of the supporting electronics and software toolchain. The paper will end with a discussion of the results of this work.

## 2. BACKGROUND

In general, attempts to collect information during the printing process can be classified by the method of data collection: either collecting information about the machine or collecting information about the printed artifact. Collecting information about the machine can be useful for determining the health of the machine and catching or predicting faults that stem from machine behavior, such as bearing wear or filament breakage. Condition based maintenance is the concept of using sensor feedback to detect developing faults in a system in order to plan and schedule maintenance only when needed and thereby minimize downtime and increase efficiency of use of machine components.

Counter to condition-based maintenance techniques, part monitoring is a method of detecting defects in the printed part by using sensors to observe the part instead of the machine. With 3D printing, these methods can be complex due to the free form nature of 3D printed parts. Research in this area has focused on a multitude of different sensor techniques, but especially machine vision applications. [8] used infrared imaging to monitor the surface temperature of the part. This information was used to derive temperature profiles for the current print layer and two layers below it, leading to information critical to interlayer bonding. [9] and [10] both utilized infrared imaging to capture spatial and temporal part temperatures. [11] mounted a camera on the print head to classify delamination faults. Pictures from the camera were processed using a neural network to successfully detect these faults. [12] used a camera to gather pictures of the corners of the printed part and then classify the deformation state using a convolutional neural network. By extracting grayscale images of the part corners, this system was able to successfully detect warping in the printed part. [13] employed a camera and image processing techniques using OpenCV to detect a number of part defects including material blobs and part detachment from the build plate. [14] applied multifractal analysis of images taken with a reflex camera to detect artifact faults in metal parts made by powder bed fusion processes. [15] used ultrasonic inspection techniques to detect delamination in solid FDM parts during manufacture.

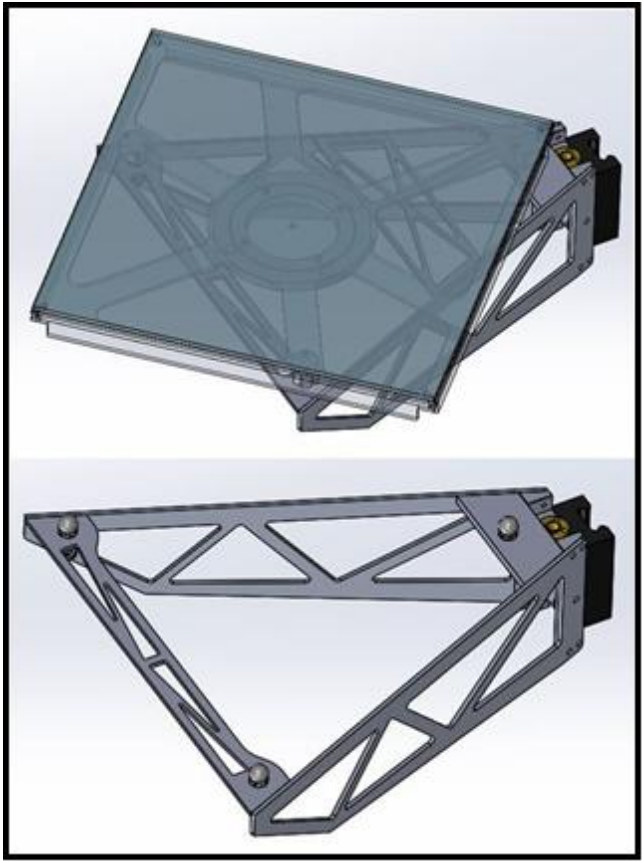

**Figure 2.** CAD renderings of the Z support frame and carriage. Note the three ball bearings forming half of the kinematic mount.

Several studies have used vision-based methods and machine learning techniques to monitor surface quality [16]–[20]. [21] and [22] explored using a single and double camera setup to compare the part profile to a profile generated from the part in a CAD program. [23] conducted a similar study but compared the pictures in a layer-wise manner. [24] utilized a multi-camera approach to monitor the entire part. This method used three pairs of cameras to create three reconstructions spaced 120 degrees apart. The reconstructions were compared with STL images of the part and were able to identify several printing errors including dimension errors, nozzle state issues, and incomplete prints.

In addition to image processing with cameras, several studies have also been undertaken using point cloud generation methods. [25] developed a framework for comparing point clouds to the STL file of a part. The STL file is used to generate a reference depth image which is compared to a depth image generated by converting a point cloud of the part. The part in this study was scanned with a laser line scanner mounted on the print head of the 3D printer. [26] proposed the use of the Fiedler number from Spectral Graph Theory as a measure of quantifying the relative quality of 3D printed surfaces. A laser line scanner was used to capture point clouds of the same part printed on two different platforms in two different materials. Use of the Fiedler number to capture differences between two parts was contrasted with statistical feature mining and facet examination techniques. [27] implemented a laser scanner on a consumer-grade polymer 3D printer and proposed a machine learning technique called self-organizing maps as a method of detecting defects using the generated point cloud. [28] and [29] utilized 3D digital image correlation to scan the part and provide feedback in near real time. [30] built a prototype machine for performing two image photogrammetry on a powder bed fusion 3D printer. [31] used photogrammetric techniques with a six-camera array to detect artifact defects after completion of the printing process. [32] also used a similar camera setup to compare printed part profiles to the same profile generated in a CAD program as a means of detecting malicious cyber-attacks to a 3D printed part. [33] wrote a review of in-situ monitoring methods for fused filament fabrication.

Kinematic couplings have been around since at least 1876 [34] and work by providing an exactly constrained interface between two parts. In three-dimensional space, everybody has six degrees of freedom. To precisely locate two parts relatives to each other, there must be exactly one constraint controlling each of those six degrees of freedom. This concept is often used in optical applications to create precision mounts for optical components. One of the common forms of the kinematic coupling are three balls mated to three V grooves. Each ball has two tangential points of contact with the V groove, one on each face. Aligning the groves at non-parallel angles to each other creates a condition where the three balls can only occupy one configuration in space if their spacing with respect to each other is held constant (as if they are all part of a single, rigid body). Since the relationship between the two bodies is exactly constrained, the interface can be separated and reassembled in a very repeatable manner. A study by [35] showed a 355 mm (14 in) diameter coupling with a repeatability of ±0.25 micron.

Markforged holds a patent [Patent US9539762B2] for using a kinematic coupling to mate the print bed of a 3D printer to a moveable stage to allow the build platform to be removed. This feature is built into all current Markforged printers [36]. Kinematic couplings have also been used for the bed of the open-source Jubilee printer [37] to provide an automated three point leveling mechanism, and on the build platform of the Fuselab FL300 printer [38].

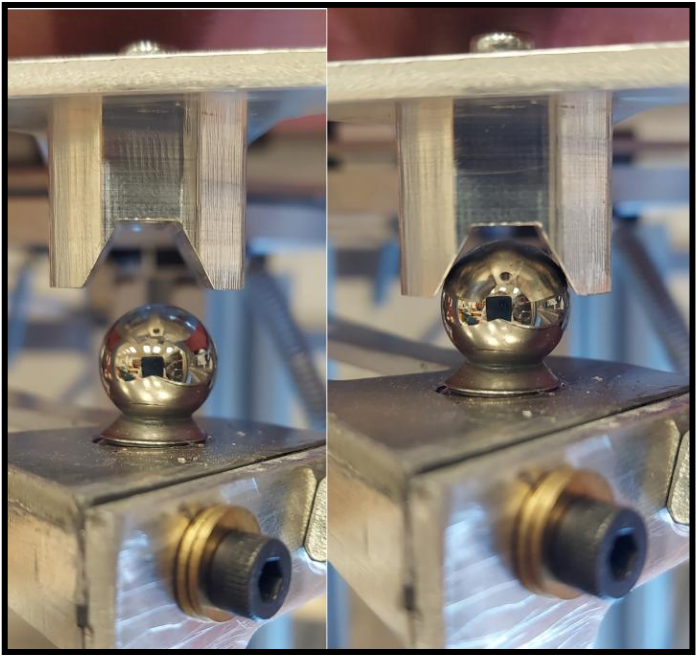

**Figure 3.** Mating of one of the ball and v-block pairs that make up the kinematic coupling.

## 3. DESIGN OF BED ROTATION MECHANISM

The bed rotation mechanism for the current platform developed by the authors consists of a kinematic coupling, a large ball bearing attached to the underside of the build plate, and support tower with stepper motor and drive shaft attached to the bottom of the enclosure. The print bed is built in typical polymer FDM style: a borosilicate build surface supported by an aluminum plate with bonded resistive heater pad. The bed is 330 mm x 330 mm of which 300 mm x 300 mm is directly over the heating coils in the heating pad. The bed is supported by a kinematic coupling consisting of three ball bearings mated to three v-blocks. The ball bearings are mounted on threaded shafts to provide the bed with a leveling mechanism, and these shafts are connected to the support frame of the Z carriage. **Figure 1** shows the print bed and Z carriage in detail. The carriage rides on a 20 mm linear rail chosen for its ability to support the moment generated by the cantilevered print bed design. The entire carriage is motivated by an acme leadscrew with 8 mm pitch. The leadscrew is not preloaded, but the weight of the Z axis ensures that there is no backlash during normal movement of the printer. The leadscrew is actuated by a NEMA 17 stepper motor.

The v-blocks that make up half of the kinematic coupling are mounted to the underside of the bed structure and are oriented so that the axis for each block parallel to the V is pointed towards the middle of the print bed. This setup provides two things: thermal growth of the build plate does not change the position of the center of the build plate, and a mating interface that allows for the bed to be removed from its support structure and then replaced with extreme repeatability.

Thermal growth is an important consideration since the bed is designed to heat up over 180°C from room temperature and to be able to cycle through this temperature span repeatedly. If the bed was rigidly constrained, thermal growth would cause the bed to warp or buckle and would not present a flat build surface, causing defects in first layer adhesion and build plate separation. If the bed were constrained on one edge and designed to allow for thermal growth, the position of the bed would shift relative to the constrained point. This means that any print algorithm that varies the temperature of the bed would move the actual position of the bed from the assumed position of the bed in the system control, making the 3D printer unsuitable as a test platform for any experimentation with fluctuating bed temperatures. A sphere in a v-block can only move parallel to the axis of the V surface, so by orienting the axis of each v-block to intersect with a point directly under the center of the print bed, the bed is designed to make the position of the center of the bed invariant with respect to bed temperature and to remain planar and stress free during thermal growth. Figure 3 shows a detailed view of the mating between the ball and v-block.

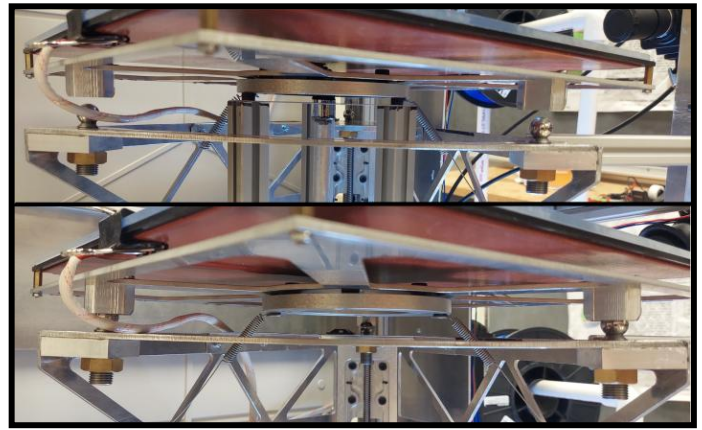

**Figure 4**. Depictions of the two states of the kinematic coupling. The top photo shows the print bed separated from the Z carriage and resting on the central support tower. The bed can be driven by the drive shaft in the support tower in this state. The bottom photo shows the print bed mated to the Z carriage and ready for printing

This kinematic coupling is fully constrained, which is to say that each degree of freedom has only a single constraining force. This means that the position of the bed is deterministic, and if the connections between the components of the coupling are broken, the bed can be replaced again in almost the exact same spot. This provides a mechanism by which the bed can be separated from the support structure of the Z motion mechanism, rotated to take pictures for photogrammetry, and then returned without disturbing the printing process. Additionally, by placing the bearing spheres on threaded shafts, the coupling also provides a method for leveling the build plate to the XY plane of the print head. A pair of springs provide a nesting force for the kinematic coupling as well as stabilizing the bed during the photogrammetry data capture cycle. These springs can be seen in the right picture of **Figure 4**.

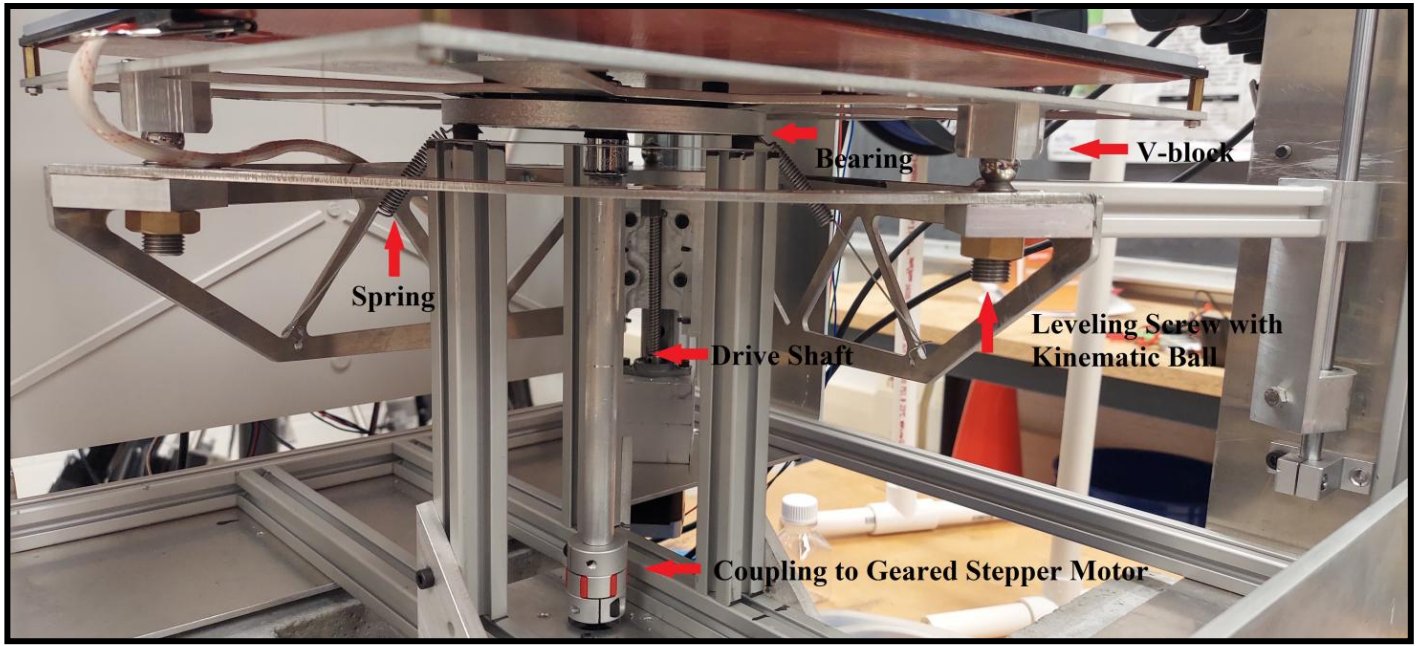

**Figure 5** Labeled depiction of the bed rotation mechanism. There are two springs to provide the nesting force and three pairs of v-blocks and leveling screws.

To rotate the bed, the inner race of a 120 mm OD diameter is affixed to the bottom of the print bed. The Z axis lowers until the outer race of this bearing comes in contact with three pillars of extruded aluminum that are connected to the bottom of the printer. These pillars then bear the weight of the print bed, and the Z carriage continues to lower until the elements of the kinematic coupling separate. Additionally, there is a hexagonal bolt head attached to the bottom center of the print bed, and when the print bed lowers onto the pillars, this bolt head is engaged by a socket that is connected by a shaft to a geared NEMA 17 stepper motor. Figure 5 shows the support structure, drive shaft, and socket. Since the bed is supported through the bearing, the stepper motor can now be used to rotate the bed. This mechanism is used to position the printed artifact for the camera so that photogrammetry can be performed. Once the capture of pictures is complete, the bed is rotated back into its original position, and the Z carriage is raised until the elements of the kinematic coupling re-engage, and the Z carriage returns the print bed to the appropriate height to resume printing. Figure 4 shows the mechanism engaged to allow bed rotation for photogrammetry data capture against the state where the kinematic coupling is engaged for printing.

## 4. ELECTRONICS AND SOFTWARE TOOLCHAIN

This process starts with the selection of the software used to perform the photogrammetry. There are several excellent programs for this application, but since open-source programs are preferred, we considered MicMac and AliceVision's Meshroom. MicMac is a command line program developed by the French National Geographic Institute and the French National School for Geographic Sciences. MicMac uses only a CPU for calculations during the photogrammetry process which means that a GPU is not needed, opening up a large choice of controllers. However, the biggest problem with MicMac is that it is not optimized to take advantage of parallelism in the photogrammetry computations and only uses a single CPU core. Due to this, dense point cloud computations can take a long time.

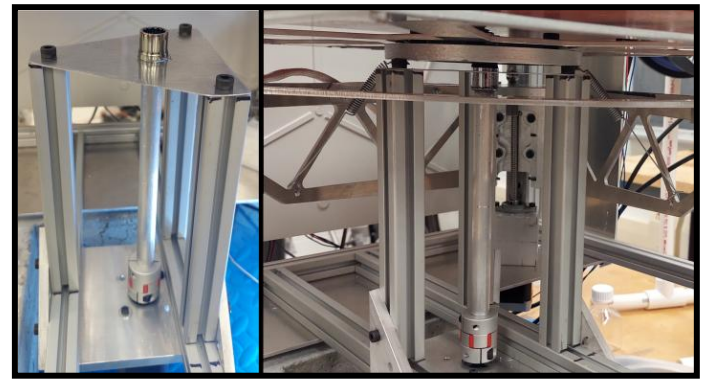

**Figure 6.** Central support tower with driveshaft for rotating the print bed. The left photograph shows the outer race of the ball bearing on the bottom of the bed resting on the central support tower.

Meshroom is a GUI built in Python to utilize the AliceVision framework for photogrammetry. Meshroom utilizes a multithreaded process that uses a GPU to perform the photogrammetry calculations that benefit from parallelism. However, the downside to this is that a CUDA enabled Nvidia GPU is required for computation. This drastically reduces the number of controllers that can be used for this system. It does significantly speed up the computation time and for this reason Meshroom was chosen as the software of choice. Ultimately, as will be explained shortly, the Meshroom GUI was not used and the AliceVision framework was utilized with a Python batch script instead.

The controller chosen for this project was a Nvidia Jetson TX2. This board is an edge computing platform that contains a dual core Denver CPU, a quad core ARM A57 CPU, a 256 core GPU, and 8Mb of RAM. While this is an amazing amount of computing power in a small package, it is not without its drawbacks. One of the biggest problems with using unusual architecture is that precompiled programs are not often available and must instead be compiled from the source code. The AliceVision framework is dependent on several other toolchains such as OpenCV, OpenImageIO, Boost, Geogram, and others which made this compilation process rather painful. Once all of these dependencies and AliceVision were built on the TX2, the 32 GB of on-board emmc memory was nearly full. Compiling Meshroom required compiling QT as a dependency, which the author was unable to accomplish. However, a batch script was written in Python to access the relevant functions of the AliceVision framework to perform photogrammetry and process a set of photographs into a dense point cloud. This method is preferable for automation since it does not need the overhead of a GUI and is a better long-term solution for this platform.

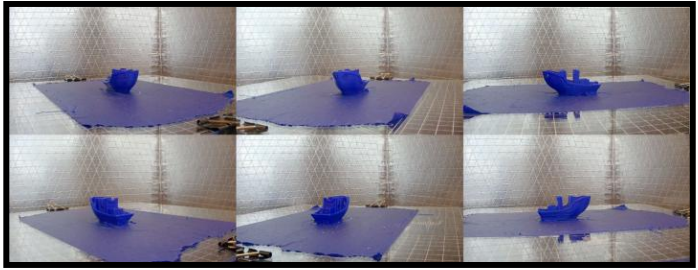

**Figure 7.** Six pictures of a Benchy captured from the lower USB camera during a scan at layer 100 of the printing process

The Jetson TX2 controls two USB web cameras to capture data for the photogrammetry process. These cameras are from Arducam and feature the 2MP AR0230 CMOS sensor. These sensors have a pixel size of 3.3 microns x 3.3 microns and a dynamic range of 105 dB. One camera is placed in the top front corner of the building chamber and the other camera placed below it near the lower limit of the print bed travel. When the bed rotation mechanism is engaged, these cameras are used to capture photographs of the printed artifact. These photographs are then processed by the Jetson TX2 into a dense point cloud which is saved for future analysis.

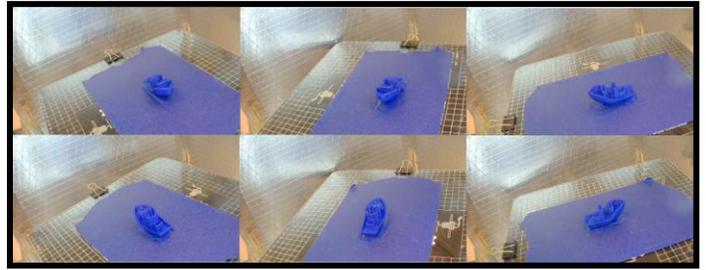

**Figure 8.** Six pictures of a Benchy captured from the upper USB camera during a scan at layer 100 of the printing process.

A G-code M word, specifically the M102 word, is used to perform the photogrammetry data capture process. The slicer program chosen for this tool chain is KISSlicer. KISSlicer allows an incredible amount of control over the slicing parameters and G code generation. One very important feature is the ability to insert custom G code every N layers. For this project, this means that every N layers the M102 word can be easily inserted into the G code to perform the photogrammetry data capture and take a record of the geometry of the printed object. The M word is passed with a P parameter that tells the printer how many positions to take pictures from during one full rotation of the bed. For instance, M102 P20 will cause the printer to take two pictures, one from each USB web camera, from 20 equally spaced positions around the object. The code for this causes the Z axis to go to its bottom position where the three vee blocks on the print bed have disengaged from the ball bearings on the motion stage. At this point, the bed is supported through the outer race of its large bearing by the three pillars mounted in the bottom of the building space and the stepper motor for controlling bed rotation is engaged with the nut on the bottom of the bed through its custom shaft and socket. The picture taking for this process is currently done manually but will be implemented to take place automatically in the future. When the KFLOP is ready to take a picture, a pop-up window is displayed on the control computer indicating that the bed is in position. The user captures a photograph from both USB cameras and then hits the "OK" button on the pop-up window and KFLOP will move the bed to the next position. Once a full rotation has been completed in this manner, the KFLOP reverses the direction of motion and returns the bed to its original position. The Z stage moves upwards, securing the print bed and then continuing to the previous Z position to reinitiate the printing process. Figures 7 through 10 show the results of this process during the printing of a Benchy test object. Figure 7 and Figure 8 were taken during the printing process, and Figure 9 and Figure 10 were taken after completion.

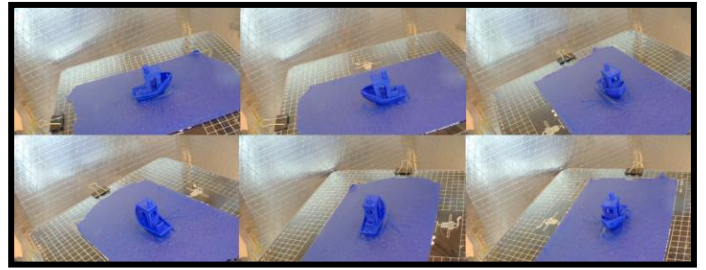

**Figure 9.** Six pictures of a Benchy captured from the upper USB camera during a scan at layer 100 of the printing process.

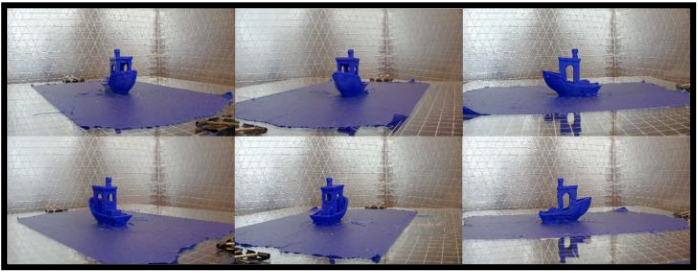

**Figure 10.** Six pictures of a Benchy captured from the upper USB camera during a scan at layer 100 of the printing process.

## 5. CONCLUSIONS & FUTURE WORK

The photogrammetry parameters need to be optimized. There are a large number of settings within the AliceVision framework that can be manipulated to produce better models from the process, and many of them drastically affect the amount of processing time it takes to turn a batch of photos into a dense point cloud. If the point cloud is to be used to provide feedback during the printing process, the photo processing time needs to be reduced to a reasonable amount of time. When the Jetson TX2 was chosen for this project, the authors felt that it represented a good amount of processing power for the task and was cheaper than other options, such as a desktop computer with graphics card. However, during testing it was realized that the amount of memory available on the system is a bottle neck for processing because the system usually runs out of memory when parallelizing some of the computation steps before it runs out of processing power. In this respect, the 8GB of memory available on the Jetson TX2 is a hinderance to processing speed. As an embedded platform, this memory is not upgradeable which presents a problem for trying to mitigate this issue in the future. Additionally, the Jetson TX2 is an edge computing platform and figuring out how to compile the AliceVision framework and working through all of the versioning problems with its dependent software packages was very time consuming. Having completed this process, the authors believe that it would be better to use a computer to do the photogrammetry processing which would open up more freedom in manipulating and using the photogrammetry process for future experiments.

This work presented a novel bed rotation mechanism that facilitates the collection of photographs of a part-in-situ during the printing process. These photographs can be processed using photogrammetric techniques to give a geometric record of the part while it is being printed. Future work for this project includes examining photogrammetric data for evidence of different defects to see if they are observable and capable of being detected automatically using methods such as point cloud comparisons. Afterwards, work can begin on correlating the captured build parameters to these detected defects.

**Acknowledgements**

The authors would like to acknowledge the support of Clemson University. All statements within are those of the authors and may or may not represent the views of these institutions.